%% file: emnlp2023.tex
\pdfoutput=1

\documentclass[11pt]{article}

\usepackage{EMNLP2023}

\usepackage{times}
\usepackage{latexsym}

\usepackage[T1]{fontenc}

\usepackage[utf8]{inputenc}

\usepackage{microtype}

\usepackage{inconsolata}

\usepackage{graphicx}
\usepackage{amsmath}
\usepackage{float}

\title{IsoChronoMeter: A simple and effective isochronic translation evaluation metric}

\author{Nikolai Rozanov$^{1,2}$~~Vikentiy Pankov$^1$~~Dmitrii Mukhutdinov$^1$~~Dima Vypirailenko$^1$\\
  $^1$Brask AI\\
    \texttt{\{vikentiy@brask.ai, dm@brask.ai, dima@brask.ai\}} \\
  $^2$Imperial College London\\
  \texttt{\{nikolai.rozanov@gmail.com\}} \\ 
  }

\begin{document}
\maketitle


\input{files/0_abstract}
\input{files/1_introduction}

\input{files/2_background}
\input{files/3_method}
\input{files/4_results}

\input{files/5_analysis}
\input{files/6_conclusion}

\bibliography{anthology,custom}
\bibliographystyle{acl_natbib}

\end{document}

%% file: files/0_abstract.tex
\begin{abstract}
Machine translation (MT) has come a long way and is readily employed in production systems to serve millions of users daily. With the recent advances in generative AI, a new form of translation is becoming possible - video dubbing. This work motivates the importance of isochronic translation, especially in the context of automatic dubbing, and introduces `IsoChronoMeter' (ICM). ICM is a simple yet effective metric to measure isochrony of translations in a scalable and resource-efficient way without the need for gold data, based on state-of-the-art text-to-speech (TTS) duration predictors. We motivate IsoChronoMeter and demonstrate its effectiveness. Using ICM we demonstrate the shortcomings of state-of-the-art translation systems and show the need for new methods. We release the code at this URL: \url{https://github.com/braskai/isochronometer}.
\end{abstract}

%% file: files/1_introduction.tex
\section{Introduction}
The isochronic translation is a practice of ensuring that the timing of speech in translated content matches the original. It has become increasingly crucial in AI-driven dubbing. As the demand for multilingual audiovisual content grows, the ability to maintain the natural rhythm and pacing of the original language through isochronic translation is vital for the success of AI dubbing systems.
Traditionally, human translators and voice actors have emphasized the importance of synchronizing translated dialogue with on-screen visuals to ensure a seamless viewing experience. This synchronization, known as isochrony, is essential for maintaining the illusion that the actors are speaking the translated language, matching their lip movements and pauses with the new audio. Recently, with the advancements in neural machine translation and text-to-speech systems, researchers have strived to replicate this isochrony automatically, aiming to preserve the speech-pause structure of the original language in the translated content \citep{tam2022isochronyawareneuralmachinetranslation, lakew2022isometricmtneuralmachine}.\\

Another way to ensure good dubbing synchronization is lip-sync. While lip-syncing is often employed to ensure synchronicity in dubbing, it presents significant challenges. Lip-syncing may force the translated dialogue to unnaturally conform to the lip movements of the original actors, potentially compromising the accuracy and fluidity of the translation. This often results in awkward or stilted dialogue, which can spoil the overall viewing experience. Additionally, due to the linguistic differences between languages, perfect lip-syncing can be impractical, leading to less faithful representations of the original content. Consequently, although lip-syncing can enhance visual alignment, it is not the optimal approach for achieving high-quality dubbing, especially when the goal is to maintain the natural flow and meaning of the original speech \citep{brannon-etal-2023-dubbing}
Research has demonstrated that integrating isochronic translation into AI dubbing significantly enhances the quality and naturalness of dubbed content, making it more acceptable to global audiences. By preserving the timing and rhythm of the original speech, these systems not only improve the technical quality of the translation but also maintain the emotional and narrative integrity of the content \citep{chronopoulou2023jointlyoptimizingtranslationsspeech}. \\

\subsection{Contribution}
In this work we present a new isochronic metric, `IsoChronoMeter' (ICM), and evaluation dataset for isochronic translation and demonstrate that `normal' translations, even by state-of-the-art systems based on LLMs and human translations, without isochrony in mind, do not achieve a good level of isochronic translation. This highlights the importance of developing specialised translation systems that are able to perform isochronic translation.

%% file: files/2_background.tex
\section{Background}
\subsection{Isochronic translation and metrics}
Initial approaches that wanted to achieve isochronic translation focused on isometric translation \citep{Federico2020, karakanta202042answersubtitlingorientedspeech, Lakew2021ISOMETRICMN, lakew2021machinetranslationverbositycontrol}, where the aim of MT systems was to translate text to achieve a similar target length. Spoken language translation benchmarks included `isometric' subtasks \citep{anastasopoulos-etal-2022-findings}. However, research showed \citep{brannon-etal-2023-dubbing} that isometric translations do not result in temporally synchronized speech after dubbing, i.e. isometricity does not really correlate with isochronicity.

This led to the most recent approaches focusing on isochronic translation instead \citep{Wu2022VideoDubberMT, Chronopoulou2023JointlyOT}. However, this direction of research is fairly new: a dedicated \emph{dubbing} task in spoken language translation benchmarks was first introduced in 2023 \citep{agrawal-etal-2023-findings}, and the degree of isochronicity is either measured subjectively by humans \citep{Federico2020} or approximated via auxiliary metrics such as phoneme-based evaluation metrics \citep{chronopoulou2023jointlyoptimizingtranslationsspeech}. VideoDubber \citep{Wu2022VideoDubberMT} was among the first to successfully employ automatic duration predictors to evaluate isochronicity of the translated text, but their `isochronic' metric is still based on human feedback, and hence cannot be applied at scale. Therefore, we conclude that there is a need to evaluate isochronic translations automatically; furthermore, since automatic dubbing pipelines in practice work with a pipeline approach (i.e. first running ASR and then later translating), it is crucial to introduce a text-based isochronic translation evaluation suite.

\subsection{Evaluation Datasets}
Collecting translation datasets requires a lot of effort especially for spoken data. Existing work includes Must-C \citep{di-gangi-etal-2019-must}, GigaST \citep{ye2023gigast10000hourpseudospeech}, CoVost-2 \citep{wang2020covost2massivelymultilingual} and Anim-400K \citep{cai2024anim400klargescaledatasetautomated}. Datasets that specificially focus on isochronic translation using professional dubbing services only seem to exist privately \citep{brannon-etal-2023-dubbing}. In our work, we choose CoVost-2 due to its permissible licenses and availability of languages.

\subsection{Identified challenges.}
A full isochrony estimation would require humans to read out the given original text and the translated text. Additionally, one would need to attempt to find speakers that have similar speaking rates in their respective languages. We propose to overcome this by a novel isochrony metric that is easy to compute without the need of human annotations (i.e. human speech) and a joined isochrony and translation quality metric without the need of gold annotations.

%% file: files/3_method.tex
\section{Method}

\subsection{Metrics}
\subsubsection{IsoChronoMeter (ours) - automatic reference-free isochrony estimation}
To compute isochrony metrics, we utilize the open-source TTSMMS project\footnote{Only a github is available: \url{https://github.com/wannaphong/ttsmms}}, which is based on Vits TTS \citep{kim2021conditionalvariationalautoencoderadversarial} and MMS \citep{pratap2023scalingspeechtechnology1000}, which supports multiple languages. Specifically, we use the duration predictor component to estimate the durations of the original text and translated texts generated by different machine translation (MT) models. As an isochrony metric, we apply a simple relative absolute error formula.
Since the duration predictors for most languages are trained on similar domains (biblical texts) and share the same architecture, we expect them to produce similar durations adjusted to the average speaking rate of each language. Therefore, we can assume that if the texts are isochronic, their durations will be close. Concretely, IsoChronoMeter (ICM) is:
\begin{align}
    &ICM =\\ \nonumber
    &\left|\left|\frac{MMS(original)-MMS(translated)}{MMS(original)}\right|\right|^2_2
\end{align}
Therefore, ICM is 0 if the duration of the original audio length prediction and the translated audio length prediction are the same; otherwise ICM represents a percentage of how much the two audio durations deviate from one another, e.g. ICM = 0.5 means that one of the audio duration predictions is half the duration prediction of the other.

\subsubsection{Blaser2.0 - automatic reference-free machine translation quality estimation}
To estimate MT quality (QE), we utilise BLASER2.0 models \citep{chen-etal-2023-blaser}, based on SONAR embeddings \citep{duquenne2023sonarsentencelevelmultimodallanguageagnostic}, to predict cross-lingual semantic similarities between the translation and original texts. Concretely,
\begin{align}
    &QE = \\ \nonumber
    &blaser2\big(sonar(original), sonar(translated)\big)
\end{align}
\citet{chen-etal-2023-blaser} show that such quality-estimation metrics outperform standard metrics such as bleu.

\subsubsection{Adjusted-IsoChronoMeter - automatic reference-free joined machine translation quality and isochrony estimation}
We also propose another metric based on the combination of IsoChronoMeter and Blaser, Adjusted-IsoChronoMeter (A-ICM). Concretely:
\begin{equation} 
AICM = (1-ICM) * QE
\end{equation}

\subsection{Effectiveness of the isochronic metric}
\begin{figure}[h]
    \centering
    \includegraphics[width=\linewidth]{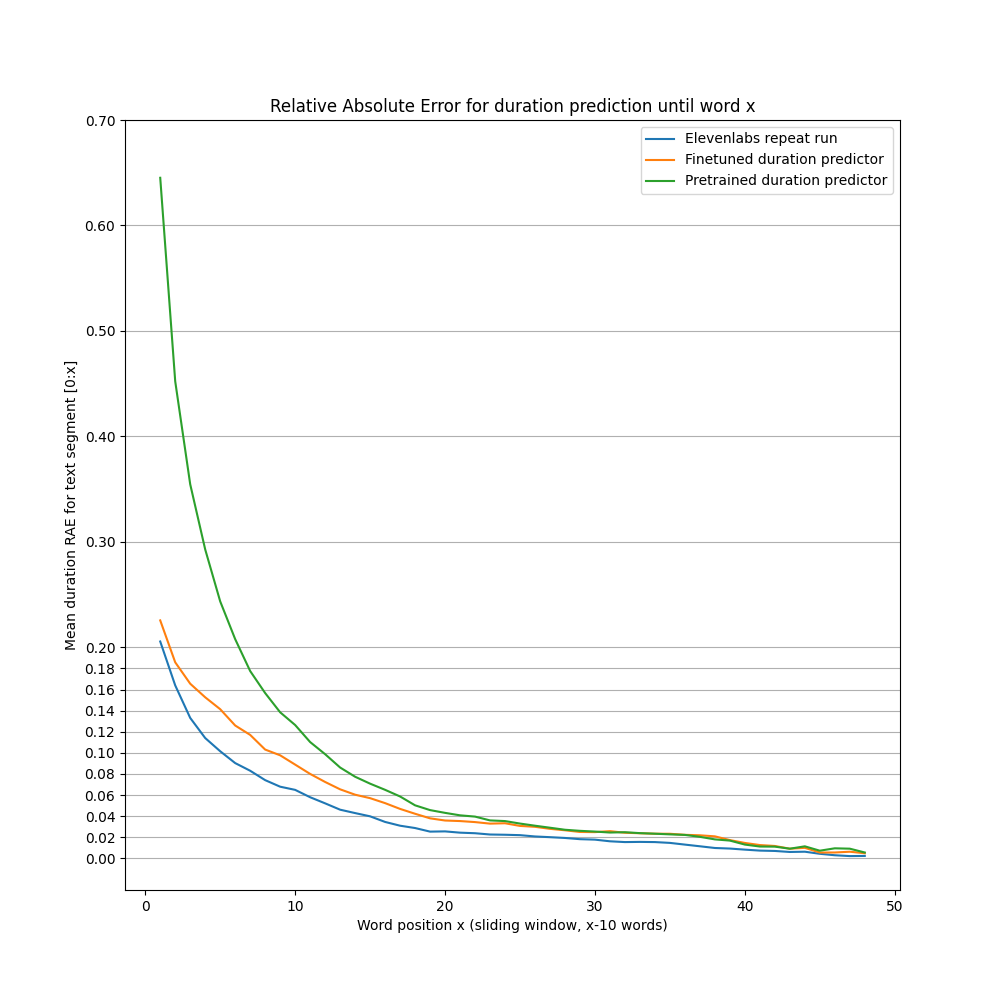}
    \caption{Dataset on English data. On the y-axis there is the relative absolute error between an original TTS-generated audio-sample and the associated prediction. On the x-axis is the number of total words used for the audio sample / prediction. Three curves show a secondary TTS-generated audio-sample (interestingly showing a big error for a few words), a fine-tuned duration predictor and the original duration predictor.}
    \label{fig:method_duration}
\end{figure}

Modern TTS systems such as Elevenlabs\footnote{\url{elevenlabs.com}} or Rask AI\footnote{\url{rask.ai}} are able to produce realistic voices and voice-clones in multiple languages. These synthetic voices share incredible similarity with human voices. Therefore, we argue that using TTS as a proxy for the duration of human speech is effective. However, since we use a duration predictor for a TTS system, we need to show that the duration predictor is faithful to the real duration of a TTS system. To show this, we conduct a simple validation against an internal dataset of a few hours of English audio data, see Figure \ref{fig:method_duration}. Concretely, for each audio file we generate the `original' TTS-generated audio sample and compare it against three predictions. Firstly, we compare against a `repeat run', i.e. we generate a second audio file using the same TTS provider. Interestingly, the repeat run does not produce 0 or close to 0 error, in fact for $<15$ words the error is above 5\%. Secondly, we compare against our standard duration predictor. Finally, we also compare against a fine-tuned version of the duration predictor. We find that for small word counts the error is quite significant for all three, but especially for the not fine-tuned duration predictor. For $x>15$, however, all three curves start converging and are within 5\% error of one another. Therefore our metric becomes effective after a sufficiently large threshold of words.
\subsection{Dataset filtering}
\begin{figure}[t]
    \centering
    \includegraphics[width=0.8\linewidth]{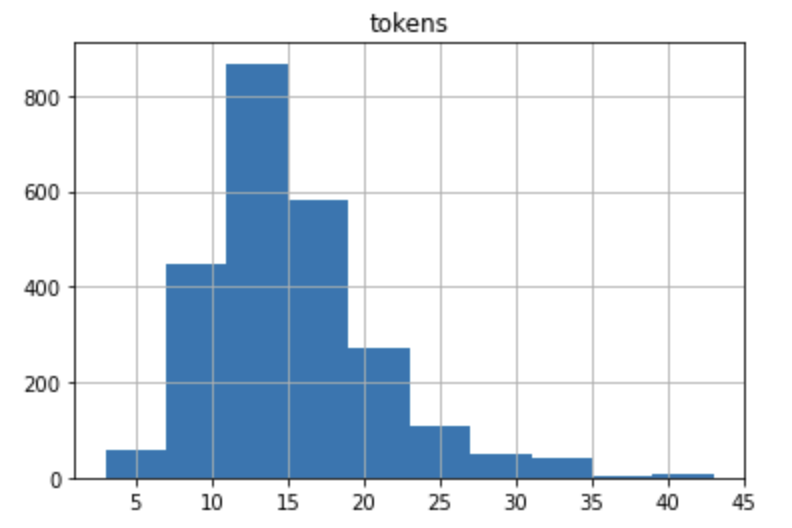}
    \caption{A histogram of sentence count vs. number of tokens in a sentence. I.e. the x-axis represents the number of tokens in a sentence, the y-axis is the total count of such sentences.}
    \label{fig:method_data}
\end{figure}

To demonstrate our metric and the need for isochronic translation engines, we create a small high-quality dataset from the CoVoST-2 \citep{wang2020covost2massivelymultilingual}, which is based on CommonVoice \citep{ardila2020commonvoicemassivelymultilingualspeech}. Specifically, taking into account the effectiveness of our metric after a specific threshold, we first filter the CoVoST-2 dataset by size. To find a good trade-off between dataset size and metric efficiency, we plot the histogram of counts and discover that above 20 tokens strikes a good balance, see Figure \ref{fig:method_data}. In particular, we observe that if we choose the number of tokens to be 25 and higher we have too few sentences, while if we choose the number of tokens to be 15 or less our duration predictor is weak, therefore 20 and above tokens is the optimal point. Additionally, we also filter the dataset based on quality rankings by humans present in the Covost dataset. Concretely, we only take data-points where there are no \texttt{downvotes} and at least three \texttt{upvotes}. The rationale behind this is to have only high quality translation samples present.

\begin{table}[H]
\centering
\begin{tabular}{|l|c|c|c|}
\hline
Model & zh-I & zh-Q & zh-A  \\
\hline
AIST-AIRC& -& -& -\\
Aya23& \textbf{0.18}& 3.96& \textbf{3.25}\\
Claude-3.5& 0.19& 3.94& 3.19\\
CommandR-plus& \textbf{0.18}& 3.93& 3.22\\
CUNI-DS& -& -& -\\
CUNI-NL& -& -& -\\
CycleL& 0.22& 2.49& 1.94\\
CycleL2& 0.39& 2.13& 1.3\\
Dubformer& -& -& -\\
Gemini-1.5-Pro& -& -& -\\
GPT-4& \textbf{0.18}& 3.98& \textbf{3.26}\\
\textbf{Human}& 0.22& 3.72& 2.9\\
HW-TSC& \textbf{0.18}& \textbf{4.01}& \textbf{3.29}\\
IKUN& 0.19& 3.84& 3.11\\
IKUN-C& 0.21& 3.76& 2.97\\
IOL\_Research& 0.19& 3.98& 3.22\\
Llama3-70B& 0.19& 3.99& 3.23\\
Mistral-Large& -& -& -\\
MSLC& -& -& -\\
NVIDIA-NeMo& 0.21& 3.9& 3.08\\
Occiglot& -& -& -\\
ONLINE-A& \textbf{0.18}& \textbf{4.03}& \textbf{3.3}\\
ONLINE-B& 0.18& 3.91& 3.21\\
ONLINE-G& 0.19& 3.91& 3.17\\
ONLINE-W& 0.18& 3.95& 3.24\\
Phi-3-Medium& -& -& -\\
TranssionMT& -& -& -\\
TSU-HITs& -& -& -\\
Unbabel-Tower70B& \textbf{0.18}& 3.95& 3.24\\
UvA-MT& 0.2& 4& 3.2\\
Yandex& -& -& -\\
ZMT& -& -& -\\
\hline
\end{tabular}
\caption{Metrics comparison across different systems. Translation from English into: zh (Chinese). Metrics correspond to: I = IsoChronoMeter ($\downarrow$), Q = Quality Estimation ($\uparrow$), A = Adjusted-IsoChronoMeter ($\uparrow$).}
\label{tbl:zh}
\end{table}

%% file: files/4_results.tex
\section{Results}
In this section we show all the results that we produce for the WMT24 shared testsuite task \citep{generalmt2024}. Specifically, all included reference paper can be found in Appendix \ref{appendix}. Our evaluation, as described above, combines three metrics: IsoChronoMeter (I), Quality Estimation (Q) and Adjusted-IsoChronoMeter (A) (see Equations (1,2,3)). In particular, we received translations with a variety of systems across four language pairs: en$\rightarrow$zh, en$\rightarrow$es, en$\rightarrow$ru, en$\rightarrow$de. In total there are four tables, one per language pair.

\begin{table}[H]
\centering
\begin{tabular}{|l|c|c|c|}
\hline
Model & es-I & es-Q & es-A  \\
\hline
AIST-AIRC& -& -& -\\
Aya23& 0.48& \textbf{4.61}& \textbf{2.4}\\
Claude-3.5& 0.5& 4.59& 2.3\\
CommandR-plus& 0.5& 4.59& 2.3\\
CUNI-DS& -& -& -\\
CUNI-NL& -& -& -\\
CycleL& 0.5& 3.52& 1.76\\
CycleL2& -& -& -\\
Dubformer& 0.47& 4.6& \textbf{2.44}\\
Gemini-1.5-Pro& -& -& -\\
GPT-4& 0.5& 4.6& 2.3\\
\textbf{Human}& 0.48& 4.42& 2.3\\
HW-TSC& -& -& -\\
IKUN& \textbf{0.46}& 4.56& \textbf{2.46}\\
IKUN-C& \textbf{0.45}& 4.5& \textbf{2.48}\\
IOL\_Research& 0.48& 4.6& 2.39\\
Llama3-70B& 0.49& \textbf{4.61}& 2.35\\
Mistral-Large& -& -& -\\
MSLC& 0.47& \textbf{4.61}& \textbf{2.44}\\
NVIDIA-NeMo& 0.47& \textbf{4.62}& \textbf{2.45}\\
Occiglot& 0.51& 4.43& 2.17\\
ONLINE-A& 0.48& 4.6& 2.39\\
ONLINE-B& 0.49& \textbf{4.64}& 2.37\\
ONLINE-G& 0.48& 4.6& 2.39\\
ONLINE-W& 0.47& 4.59& 2.43\\
Phi-3-Medium& -& -& -\\
TranssionMT& 0.5& \textbf{4.62}& 2.31\\
TSU-HITs& 0.25& 3.39& 2.54\\
Unbabel-Tower70B& 0.5& \textbf{4.62}& 2.31\\
UvA-MT& -& -& -\\
Yandex& -& -& -\\
ZMT& 0.49& \textbf{4.61}& 2.35\\
\hline
\end{tabular}
\caption{Metrics comparison across different systems. Translation from English into: es (Spanish). Metrics correspond to: I = IsoChronoMeter ($\downarrow$), Q = Quality Estimation ($\uparrow$), A = Adjusted-IsoChronoMeter ($\uparrow$).}
\label{tbl:es}
\end{table}

\begin{table}[H]
\centering
\begin{tabular}{|l|c|c|c|}
\hline
Model & ru-I & ru-Q & ru-A  \\
\hline
AIST-AIRC& -& -& -\\
Aya23& 0.48& 4.91& 2.55\\
Claude-3.5& 0.49& \textbf{4.95}& 2.52\\
CommandR-plus& 0.49& 4.9& 2.5\\
CUNI-DS& 0.47& 4.86& 2.58\\
CUNI-NL& -& -& -\\
CycleL& 0.39& 3.15& 1.92\\
CycleL2& 0.3& 2.52& 1.76\\
Dubformer& \textbf{0.42}& 4.82& \textbf{2.8}\\
Gemini-1.5-Pro& -& -& -\\
GPT-4& 0.47& 4.93& 2.61\\
\textbf{Human}& 0.53& 4.82& 2.27\\
HW-TSC& -& -& -\\
IKUN& \textbf{0.42}& 4.84& \textbf{2.81}\\
IKUN-C& \textbf{0.41}& 4.77& \textbf{2.81}\\
IOL\_Research& 0.47& 4.93& 2.61\\
Llama3-70B& 0.49& \textbf{4.94}& 2.52\\
Mistral-Large& -& -& -\\
MSLC& -& -& -\\
NVIDIA-NeMo& 0.48& 4.93& 2.56\\
Occiglot& -& -& -\\
ONLINE-A& 0.48& 4.91& 2.55\\
ONLINE-B& 0.47& 4.93& 2.61\\
ONLINE-G& 0.51& 4.93& 2.42\\
ONLINE-W& 0.47& 4.92& 2.61\\
Phi-3-Medium& -& -& -\\
TranssionMT& 0.47& 4.93& 2.61\\
TSU-HITs& 0.34& 3.66& 2.42\\
Unbabel-Tower70B& 0.49& 4.92& 2.51\\
UvA-MT& -& -& -\\
Yandex& 0.48& 4.83& 2.51\\
ZMT& 0.48& 4.91& 2.55\\
\hline
\end{tabular}
\caption{Metrics comparison across different systems. Translation from English into: ru (Russian). Metrics correspond to: I = IsoChronoMeter ($\downarrow$), Q = Quality Estimation ($\uparrow$), A = Adjusted-IsoChronoMeter ($\uparrow$).}
\label{tbl:ru}
\end{table}

%% file: files/5_analysis.tex
\section{Findings}
We identify three main findings. Firstly, isochrony is not the natural way of translation (even for humans). Secondly, systems designed for dubbing, such as DubFormer, or multi-linguality, such as Aya23, outperform their `standard' counter-parts. Finally, the metric itself is powerful and determines systems that are better at dubbing without gold annotations.
\subsection{Isochrony does not come automatically}
We discover that across all language pairs, the smallest isochronic score (ICM) that we discover is 0.18, which means that the translated audio duration prediction is almost 18\% longer or shorter than the original audio prediction.

\begin{table}[H]
\centering
\begin{tabular}{|l|c|c|c|}
\hline
Model & de-I & de-Q & de-A  \\
\hline
AIST-AIRC& \textbf{0.35}& 4.69& \textbf{3.05}\\
Aya23& 0.38& 4.68& 2.9\\
Claude-3.5& 0.39& \textbf{4.7}& 2.87\\
CommandR-plus& 0.38& 4.68& 2.9\\
CUNI-DS& -& -& -\\
CUNI-NL& \textbf{0.33}& 4.62& \textbf{3.1}\\
CycleL& 0.4& 3.65& 2.19\\
CycleL2& 0.4& 3.65& 2.19\\
Dubformer& \textbf{0.32}& 4.51& \textbf{3.07}\\
Gemini-1.5-Pro& -& -& -\\
GPT-4& 0.39& \textbf{4.72}& 2.88\\
\textbf{Human}& 0.38& 4.47& 2.77\\
HW-TSC& -& -& -\\
IKUN& \textbf{0.34}& 4.57& \textbf{3.02}\\
IKUN-C& \textbf{0.34}& 4.5& 2.97\\
IOL\_Research& 0.37& \textbf{4.7}& 2.96\\
Llama3-70B& 0.39& \textbf{4.73}& 2.89\\
Mistral-Large& -& -& -\\
MSLC& 0.36& 4.61& 2.95\\
NVIDIA-NeMo& 0.37& \textbf{4.72}& 2.97\\
Occiglot& 0.46& 4.55& 2.46\\
ONLINE-A& 0.37& 4.68& 2.95\\
ONLINE-B& 0.37& 4.6& 2.9\\
ONLINE-G& 0.36& 4.69& \textbf{3}\\
ONLINE-W& 0.37& 4.66& 2.94\\
Phi-3-Medium& -& -& -\\
TranssionMT& 0.37& 4.6& 2.9\\
TSU-HITs& 0.34& 3.37& 2.22\\
Unbabel-Tower70B& 0.38& 4.68& 2.9\\
UvA-MT& -& -& -\\
Yandex& -& -& -\\
ZMT& 0.37& 4.68& 2.95\\
\hline
\end{tabular}
\caption{Metrics comparison across different systems. Translation from English into: de (German). Metrics correspond to: I = IsoChronoMeter ($\downarrow$), Q = Quality Estimation ($\uparrow$), A = Adjusted-IsoChronoMeter ($\uparrow$).}
\label{tbl:de}
\end{table}

\subsection{Most promising systems}
The most promising systems that are overall better at isochronic translation as well as translation quality are DubFormer, Ikun, Ikun-C \citep{wmt24_id19} and Cuni-NL \citep{wmt24_id16}. For some language pairs, some big players such as GPT-4, Nemo and `Online A' perform well as well as some specialised systems HW-TSC \citep{wmt24_id4} and MSLC \citep{wmt24_id2}. Aya23 outperforms its backbone model CommandR-plus, which is intuititve and show that multi-linguality helps MT and isochronic-MT.
\subsection{Nuances in the metric}
Overall we discover that the joined metric is very powerful in ranking systems. We discover an edge case for en$\rightarrow$zh, where TSU-HITs has a poor translation quality and likely drops parts of the translation, resulting in poor quality estimate scores, but it has excellent isochrony scores and adjusted isochrony scores. Therefore, we recommend using a performance threshold when applying the metric.


%% file: files/6_conclusion.tex
\section{Conclusion \& Future Work}
We motivate the importance of isochronic translation. To this end, we presented a novel and simple metric to evaluate isochrony that does not require gold annotations. We evaluated the shared task and discovered that: 1. Isochrony does not come naturally for translation systems, including human (non-isochronic) translation; 2. Systems and LLMs that are designed for multi-linguality or dubbing perform better on our main metric `Ajusted-IsoChronoMeter', which combines isochrony and machine translation quality; 3. The metric requires some nuance, as systems that drop parts of the translation might have a good isochrony score, but bad translation quality score - overall biasing them towards a better A-ICM. 

\subsection{Future directions}
There are several future directions that we identify. Firstly, isochronic translation itself is a promising direction and automatic metrics such as IsoChronoMeter can help with advancing this field. Secondly, extending the benchmark to include gold human translation designed for dubbing. Finally, a more detailed evaluation and improvement of the metric itself; specifically, we believe better duration predictors are possible, and more rigorous evaluation, including using gold annotations and on more language pairs.

\subsection{Acknowledgements}
We want to thank all our colleagues at Brask and Rask AI that have supported this work and their generous contributions in time and resources to facilitate this work. Likewise, we thank our colleagues at Imperial College London. We also want to thank the WMT organisers, especially Eleftherios Avramidis and Tom Kocmi.

%% file: emnlp2023.bbl
\begin{thebibliography}{26}
\expandafter\ifx\csname natexlab\endcsname\relax\def\natexlab#1{#1}\fi

\bibitem[{Agarwal et~al.(2023)Agarwal, Agrawal, Anastasopoulos, Bentivogli, Bojar, Borg, Carpuat, Cattoni, Cettolo, Chen, Chen, Choukri, Chronopoulou, Currey, Declerck, Dong, Duh, Est{\`e}ve, Federico, Gahbiche, Haddow, Hsu, Mon~Htut, Inaguma, Javorsk{\'y}, Judge, Kano, Ko, Kumar, Li, Ma, Mathur, Matusov, McNamee, P.~McCrae, Murray, Nadejde, Nakamura, Negri, Nguyen, Niehues, Niu, Kr.~Ojha, E.~Ortega, Pal, Pino, van~der Plas, Pol{\'a}k, Rippeth, Salesky, Shi, Sperber, St{\"u}ker, Sudoh, Tang, Thompson, Tran, Turchi, Waibel, Wang, Watanabe, and Zevallos}]{agrawal-etal-2023-findings}
Milind Agarwal, Sweta Agrawal, Antonios Anastasopoulos, Luisa Bentivogli, Ond{\v{r}}ej Bojar, Claudia Borg, Marine Carpuat, Roldano Cattoni, Mauro Cettolo, Mingda Chen, William Chen, Khalid Choukri, Alexandra Chronopoulou, Anna Currey, Thierry Declerck, Qianqian Dong, Kevin Duh, Yannick Est{\`e}ve, Marcello Federico, Souhir Gahbiche, Barry Haddow, Benjamin Hsu, Phu Mon~Htut, Hirofumi Inaguma, D{\'a}vid Javorsk{\'y}, John Judge, Yasumasa Kano, Tom Ko, Rishu Kumar, Pengwei Li, Xutai Ma, Prashant Mathur, Evgeny Matusov, Paul McNamee, John P.~McCrae, Kenton Murray, Maria Nadejde, Satoshi Nakamura, Matteo Negri, Ha~Nguyen, Jan Niehues, Xing Niu, Atul Kr.~Ojha, John E.~Ortega, Proyag Pal, Juan Pino, Lonneke van~der Plas, Peter Pol{\'a}k, Elijah Rippeth, Elizabeth Salesky, Jiatong Shi, Matthias Sperber, Sebastian St{\"u}ker, Katsuhito Sudoh, Yun Tang, Brian Thompson, Kevin Tran, Marco Turchi, Alex Waibel, Mingxuan Wang, Shinji Watanabe, and Rodolfo Zevallos. 2023.
\newblock \href {https://doi.org/10.18653/v1/2023.iwslt-1.1} {{FINDINGS} {OF} {THE} {IWSLT} 2023 {EVALUATION} {CAMPAIGN}}.
\newblock In \emph{Proceedings of the 20th International Conference on Spoken Language Translation (IWSLT 2023)}, pages 1--61, Toronto, Canada (in-person and online). Association for Computational Linguistics.

\bibitem[{Anastasopoulos et~al.(2022)Anastasopoulos, Barrault, Bentivogli, Zanon~Boito, Bojar, Cattoni, Currey, Dinu, Duh, Elbayad, Emmanuel, Est{\`e}ve, Federico, Federmann, Gahbiche, Gong, Grundkiewicz, Haddow, Hsu, Javorsk{\'y}, Kloudov{\'a}, Lakew, Ma, Mathur, McNamee, Murray, N{\v{a}}dejde, Nakamura, Negri, Niehues, Niu, Ortega, Pino, Salesky, Shi, Sperber, St{\"u}ker, Sudoh, Turchi, Virkar, Waibel, Wang, and Watanabe}]{anastasopoulos-etal-2022-findings}
Antonios Anastasopoulos, Lo{\"\i}c Barrault, Luisa Bentivogli, Marcely Zanon~Boito, Ond{\v{r}}ej Bojar, Roldano Cattoni, Anna Currey, Georgiana Dinu, Kevin Duh, Maha Elbayad, Clara Emmanuel, Yannick Est{\`e}ve, Marcello Federico, Christian Federmann, Souhir Gahbiche, Hongyu Gong, Roman Grundkiewicz, Barry Haddow, Benjamin Hsu, D{\'a}vid Javorsk{\'y}, V{\u{e}}ra Kloudov{\'a}, Surafel Lakew, Xutai Ma, Prashant Mathur, Paul McNamee, Kenton Murray, Maria N{\v{a}}dejde, Satoshi Nakamura, Matteo Negri, Jan Niehues, Xing Niu, John Ortega, Juan Pino, Elizabeth Salesky, Jiatong Shi, Matthias Sperber, Sebastian St{\"u}ker, Katsuhito Sudoh, Marco Turchi, Yogesh Virkar, Alexander Waibel, Changhan Wang, and Shinji Watanabe. 2022.
\newblock \href {https://doi.org/10.18653/v1/2022.iwslt-1.10} {Findings of the {IWSLT} 2022 evaluation campaign}.
\newblock In \emph{Proceedings of the 19th International Conference on Spoken Language Translation (IWSLT 2022)}, pages 98--157, Dublin, Ireland (in-person and online). Association for Computational Linguistics.

\bibitem[{Ardila et~al.(2020)Ardila, Branson, Davis, Henretty, Kohler, Meyer, Morais, Saunders, Tyers, and Weber}]{ardila2020commonvoicemassivelymultilingualspeech}
Rosana Ardila, Megan Branson, Kelly Davis, Michael Henretty, Michael Kohler, Josh Meyer, Reuben Morais, Lindsay Saunders, Francis~M. Tyers, and Gregor Weber. 2020.
\newblock \href {http://arxiv.org/abs/1912.06670} {Common voice: A massively-multilingual speech corpus}.

\bibitem[{Brannon et~al.(2023)Brannon, Virkar, and Thompson}]{brannon-etal-2023-dubbing}
William Brannon, Yogesh Virkar, and Brian Thompson. 2023.
\newblock \href {https://doi.org/10.1162/tacl_a_00551} {Dubbing in practice: A large scale study of human localization with insights for automatic dubbing}.
\newblock \emph{Transactions of the Association for Computational Linguistics}, 11:419--435.

\bibitem[{Cai et~al.(2024)Cai, Liu, and Chan}]{cai2024anim400klargescaledatasetautomated}
Kevin Cai, Chonghua Liu, and David~M. Chan. 2024.
\newblock \href {http://arxiv.org/abs/2401.05314} {Anim-400k: A large-scale dataset for automated end-to-end dubbing of video}.

\bibitem[{Chen et~al.(2023)Chen, Duquenne, Andrews, Kao, Mourachko, Schwenk, and Costa-juss{\`a}}]{chen-etal-2023-blaser}
Mingda Chen, Paul-Ambroise Duquenne, Pierre Andrews, Justine Kao, Alexandre Mourachko, Holger Schwenk, and Marta~R. Costa-juss{\`a}. 2023.
\newblock \href {https://doi.org/10.18653/v1/2023.acl-long.504} {{BLASER}: A text-free speech-to-speech translation evaluation metric}.
\newblock In \emph{Proceedings of the 61st Annual Meeting of the Association for Computational Linguistics (Volume 1: Long Papers)}, pages 9064--9079, Toronto, Canada. Association for Computational Linguistics.

\bibitem[{Chronopoulou et~al.(2023{\natexlab{a}})Chronopoulou, Thompson, Mathur, Virkar, Lakew, and Federico}]{chronopoulou2023jointlyoptimizingtranslationsspeech}
Alexandra Chronopoulou, Brian Thompson, Prashant Mathur, Yogesh Virkar, Surafel~M. Lakew, and Marcello Federico. 2023{\natexlab{a}}.
\newblock \href {http://arxiv.org/abs/2302.12979} {Jointly optimizing translations and speech timing to improve isochrony in automatic dubbing}.

\bibitem[{Chronopoulou et~al.(2023{\natexlab{b}})Chronopoulou, Thompson, Mathur, Virkar, Lakew, and Federico}]{Chronopoulou2023JointlyOT}
Alexandra Chronopoulou, Brian Thompson, Prashant Mathur, Yogesh Virkar, Surafel~Melaku Lakew, and Marcello Federico. 2023{\natexlab{b}}.
\newblock \href {https://api.semanticscholar.org/CorpusID:257219873} {Jointly optimizing translations and speech timing to improve isochrony in automatic dubbing}.
\newblock \emph{ArXiv}, abs/2302.12979.

\bibitem[{Di~Gangi et~al.(2019)Di~Gangi, Cattoni, Bentivogli, Negri, and Turchi}]{di-gangi-etal-2019-must}
Mattia~A. Di~Gangi, Roldano Cattoni, Luisa Bentivogli, Matteo Negri, and Marco Turchi. 2019.
\newblock \href {https://doi.org/10.18653/v1/N19-1202} {{M}u{ST}-{C}: a {M}ultilingual {S}peech {T}ranslation {C}orpus}.
\newblock In \emph{Proceedings of the 2019 Conference of the North {A}merican Chapter of the Association for Computational Linguistics: Human Language Technologies, Volume 1 (Long and Short Papers)}, pages 2012--2017, Minneapolis, Minnesota. Association for Computational Linguistics.

\bibitem[{Duquenne et~al.(2023)Duquenne, Schwenk, and Sagot}]{duquenne2023sonarsentencelevelmultimodallanguageagnostic}
Paul-Ambroise Duquenne, Holger Schwenk, and Benoît Sagot. 2023.
\newblock \href {http://arxiv.org/abs/2308.11466} {Sonar: Sentence-level multimodal and language-agnostic representations}.

\bibitem[{Federico et~al.(2020)Federico, Enyedi, Barra-Chicote, Giri, Isik, Krishnaswamy, and Sawaf}]{Federico2020}
Marcello Federico, Robert Enyedi, Roberto Barra-Chicote, Ritwik Giri, Umut Isik, Arvindh Krishnaswamy, and Hassan Sawaf. 2020.
\newblock \href {https://www.amazon.science/publications/from-speech-to-speech-translation-to-automatic-dubbing} {From speech-to-speech translation to automatic dubbing}.
\newblock In \emph{IWSLT 2020}.

\bibitem[{Hrabal et~al.(2024)Hrabal, Jon, Popel, Luu, Semin, and Bojar}]{wmt24_id16}
Miroslav Hrabal, Josef Jon, Martin Popel, Nam Luu, Danil Semin, and Ond{\v{r}}ej Bojar. 2024.
\newblock {CUNI} at {WMT24} general translation task: Llms, (q)lora, {CPO} and model merging.
\newblock In \emph{Proceedings of the Ninth Conference on Machine Translation}, USA. Association for Computational Linguistics.

\bibitem[{Karakanta et~al.(2020)Karakanta, Negri, and Turchi}]{karakanta202042answersubtitlingorientedspeech}
Alina Karakanta, Matteo Negri, and Marco Turchi. 2020.
\newblock \href {http://arxiv.org/abs/2006.01080} {Is 42 the answer to everything in subtitling-oriented speech translation?}

\bibitem[{Kim et~al.(2021)Kim, Kong, and Son}]{kim2021conditionalvariationalautoencoderadversarial}
Jaehyeon Kim, Jungil Kong, and Juhee Son. 2021.
\newblock \href {http://arxiv.org/abs/2106.06103} {Conditional variational autoencoder with adversarial learning for end-to-end text-to-speech}.

\bibitem[{Kocmi et~al.(2024)Kocmi, Avramidis, Bawden, Bojar, Dvorkovich, Federmann, Fishel, Freitag, Gowda, Grundkiewicz, Haddow, Karpinska, Koehn, Marie, Monz, Murray, Nagata, Popel, Popovi{\'c}, Shmatova, Steingrímsson, and Zouhar}]{generalmt2024}
Tom Kocmi, Eleftherios Avramidis, Rachel Bawden, Ond{\v{r}}ej Bojar, Anton Dvorkovich, Christian Federmann, Mark Fishel, Markus Freitag, Thamme Gowda, Roman Grundkiewicz, Barry Haddow, Marzena Karpinska, Philipp Koehn, Benjamin Marie, Christof Monz, Kenton Murray, Masaaki Nagata, Martin Popel, Maja Popovi{\'c}, Mariya Shmatova, Steinþór Steingrímsson, and Vil\'{e}m Zouhar. 2024.
\newblock Findings of the {WMT}24 general machine translation shared task:\\the {LLM} era is here but mt is not solved yet.
\newblock In \emph{Proceedings of the Ninth Conference on Machine Translation}, USA. Association for Computational Linguistics.

\bibitem[{Lakew et~al.(2021{\natexlab{a}})Lakew, Federico, Wang, Hoang, Virkar, Barra-Chicote, and Enyedi}]{lakew2021machinetranslationverbositycontrol}
Surafel~M. Lakew, Marcello Federico, Yue Wang, Cuong Hoang, Yogesh Virkar, Roberto Barra-Chicote, and Robert Enyedi. 2021{\natexlab{a}}.
\newblock \href {http://arxiv.org/abs/2110.03847} {Machine translation verbosity control for automatic dubbing}.

\bibitem[{Lakew et~al.(2022)Lakew, Virkar, Mathur, and Federico}]{lakew2022isometricmtneuralmachine}
Surafel~M. Lakew, Yogesh Virkar, Prashant Mathur, and Marcello Federico. 2022.
\newblock \href {http://arxiv.org/abs/2112.08682} {Isometric mt: Neural machine translation for automatic dubbing}.

\bibitem[{Lakew et~al.(2021{\natexlab{b}})Lakew, Virkar, Mathur, and Federico}]{Lakew2021ISOMETRICMN}
Surafel~Melaku Lakew, Yogesh Virkar, Prashant Mathur, and Marcello Federico. 2021{\natexlab{b}}.
\newblock \href {https://api.semanticscholar.org/CorpusID:245219201} {Isometric mt: Neural machine translation for automatic dubbing}.
\newblock \emph{ICASSP 2022 - 2022 IEEE International Conference on Acoustics, Speech and Signal Processing (ICASSP)}, pages 6242--6246.

\bibitem[{Larkin et~al.(2024)Larkin, Lo, and Knowles}]{wmt24_id2}
Samuel Larkin, Chi-kiu Lo, and Rebecca Knowles. 2024.
\newblock {MSLC24} submissions to the general machine translation task.
\newblock In \emph{Proceedings of the Ninth Conference on Machine Translation}, USA. Association for Computational Linguistics.

\bibitem[{Liao et~al.(2024)Liao, Herold, Khadivi, and Monz}]{wmt24_id19}
Baohao Liao, Christian Herold, Shahram Khadivi, and Christof Monz. 2024.
\newblock {IKUN} for {WMT24} general {MT} task: Llms are here for multilingual machine translation.
\newblock In \emph{Proceedings of the Ninth Conference on Machine Translation}, USA. Association for Computational Linguistics.

\bibitem[{Pratap et~al.(2023)Pratap, Tjandra, Shi, Tomasello, Babu, Kundu, Elkahky, Ni, Vyas, Fazel-Zarandi, Baevski, Adi, Zhang, Hsu, Conneau, and Auli}]{pratap2023scalingspeechtechnology1000}
Vineel Pratap, Andros Tjandra, Bowen Shi, Paden Tomasello, Arun Babu, Sayani Kundu, Ali Elkahky, Zhaoheng Ni, Apoorv Vyas, Maryam Fazel-Zarandi, Alexei Baevski, Yossi Adi, Xiaohui Zhang, Wei-Ning Hsu, Alexis Conneau, and Michael Auli. 2023.
\newblock \href {http://arxiv.org/abs/2305.13516} {Scaling speech technology to 1,000+ languages}.

\bibitem[{Tam et~al.(2022)Tam, Lakew, Virkar, Mathur, and Federico}]{tam2022isochronyawareneuralmachinetranslation}
Derek Tam, Surafel~M. Lakew, Yogesh Virkar, Prashant Mathur, and Marcello Federico. 2022.
\newblock \href {http://arxiv.org/abs/2112.08548} {Isochrony-aware neural machine translation for automatic dubbing}.

\bibitem[{Wang et~al.(2020)Wang, Wu, and Pino}]{wang2020covost2massivelymultilingual}
Changhan Wang, Anne Wu, and Juan Pino. 2020.
\newblock \href {http://arxiv.org/abs/2007.10310} {Covost 2 and massively multilingual speech-to-text translation}.

\bibitem[{Wu et~al.(2022)Wu, Guo, Tan, Zhang, Li, Song, He, Zhao, Menezes, and Bian}]{Wu2022VideoDubberMT}
Yihan Wu, Junliang Guo, Xuejiao Tan, Chen Zhang, Bohan Li, Ruihua Song, Lei He, Sheng Zhao, Arul Menezes, and Jiang Bian. 2022.
\newblock \href {https://api.semanticscholar.org/CorpusID:254095829} {Videodubber: Machine translation with speech-aware length control for video dubbing}.
\newblock \emph{ArXiv}, abs/2211.16934.

\bibitem[{Wu et~al.(2024)Wu, Wei, Li, Shang, GUO, Li, Rao, Luo, Xie, and Yang}]{wmt24_id4}
Zhanglin Wu, Daimeng Wei, Zongyao Li, Hengchao Shang, Jiaxin GUO, Shaojun Li, Zhiqiang Rao, Yuanchang Luo, Ning Xie, and Hao Yang. 2024.
\newblock Choose the final translation from {NMT} and {LLM} hypotheses using {MBR} decoding: {HW}-{TSC}'s submission to the {WMT24} general {MT} shared task.
\newblock In \emph{Proceedings of the Ninth Conference on Machine Translation}, USA. Association for Computational Linguistics.

\bibitem[{Ye et~al.(2023)Ye, Zhao, Ko, Meng, Wang, Wang, and Cao}]{ye2023gigast10000hourpseudospeech}
Rong Ye, Chengqi Zhao, Tom Ko, Chutong Meng, Tao Wang, Mingxuan Wang, and Jun Cao. 2023.
\newblock \href {http://arxiv.org/abs/2204.03939} {Gigast: A 10,000-hour pseudo speech translation corpus}.

\end{thebibliography}
